\documentclass{article}

\usepackage{url}
\usepackage{times}
\usepackage{amsmath}
\usepackage{graphicx}
\usepackage{import}
\usepackage{subfig}
\usepackage{leadsheets}
\usepackage[vlined,algoruled,linesnumbered]{algorithm2e}
\usepackage{todonotes}
\usepackage{multicol}

\newcommand{\ie}{i.e.,\,}
\newcommand{\eg}{e.g.,\,}
\newcommand{\len}{\ensuremath{n}}

\renewcommand{\song}[1]{\emph{``#1''}}
\newcommand{\iasm}{\song{In a Sentimental Mood}}
\newcommand{\ch}[1]{\text{\writechord{#1}}}

\newcommand{\note}[2]{\ensuremath{\text{#1}_{#2}}}

\title{Sampling Variations of Lead Sheets}
\author{Pierre Roy, Alexandre Papadopoulos, Fran\c{c}ois Pachet\\Sony CSL, Paris\\roypie@gmail.com, pachetcsl@gmail.com, alexandre.papadopoulos@lip6.fr}

\begin{document}

\maketitle

\begin{abstract}
Machine-learning techniques have been recently used with spectacular results to generate artefacts such as music or text. However, these techniques are still unable to capture and generate artefacts that are convincingly structured. In this paper we present an approach to generate structured musical sequences. We introduce a mechanism for sampling efficiently variations of musical sequences. Given a input sequence and a statistical model, this mechanism samples a set of sequences whose distance to the input sequence is approximately within specified bounds. This mechanism is implemented as an extension of belief propagation, and uses local fields to bias the generation. We show experimentally that sampled sequences are indeed closely correlated to the standard musical similarity measure defined by Mongeau and Sankoff. We then show how this mechanism can used to implement composition strategies that enforce arbitrary structure on a musical lead sheet generation problem.\end{abstract}

\section{Introduction}

Recent advances in machine learning, especially deep recurrent networks such as Long short-term memeory networks (LSTMs), led to major improvements in the quality of music generation \cite{hadjeres:16a,Lyu:2015:MHS:2832747.2832827}. They achieve spectacular performance for short musical fragments. However, musical structure typically exceeds the scope of statistical models. As a consequence the resulting music
lacks a sense of direction and becomes boring to the listener after a short while~\cite{waite_2016}.

Musical structure is the overall organisation of a composition into sections, phrases, and patterns, very much like the organisation of a text.
The structure of musical pieces is scarcely, if ever, linear as it essentially relies on the repetition of these elements, possibly altered. For example, songs are decomposed into repeating sections, called verses and choruses, and each section is constructed with repeating patterns. In fact, the striking illusion discovered by \cite{deutsch2011illusory} shows that repetition truly \emph{creates} music, for instance by turning speech into music. This is further confirmed by \cite{margulis2013aesthetic} who observed that inserting arbitrary repetition in non-repetitive music improves listeners rating and confidence that the music was written by a human composer.

\begin{figure}[htp]
	\centering
		\includegraphics[width=0.95\columnwidth]{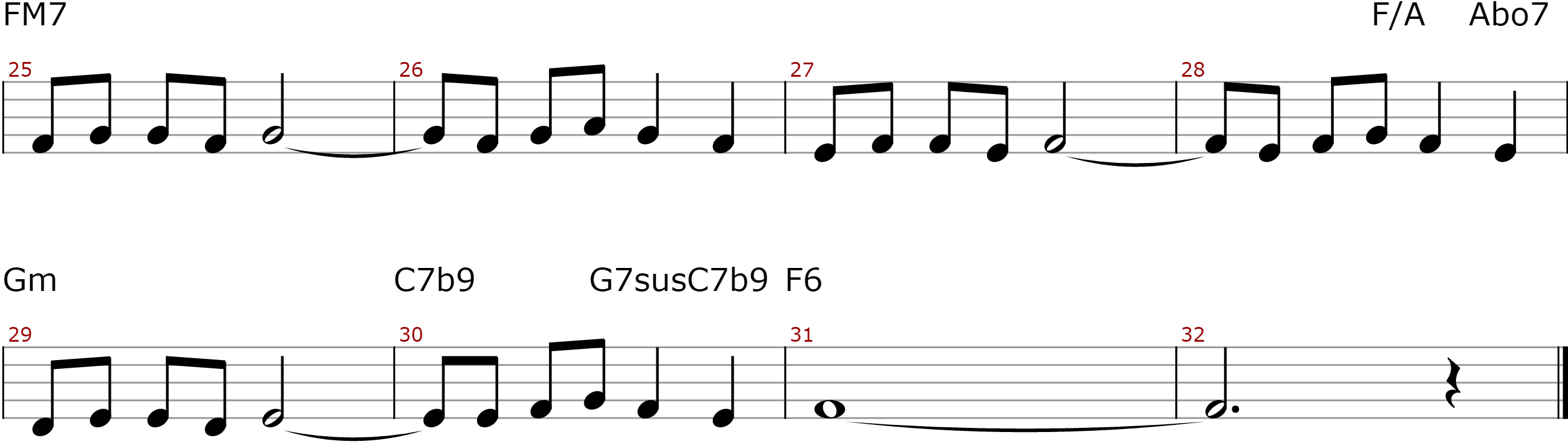}
	\caption{The last eight bars of \song{Strangers in the Night}.}
	\label{fig:strangers}
\end{figure}

\emph{Variations} are a specific type of repetition, in which the original melody is altered in its rhythm, pitch sequence, and/or harmony. Variations are used to create diversity and surprise by subtle, unexpected changes in a repetition.
%
The song \song{Strangers in the Night} is a typical 32-bar form with an AABA structure consisting of four 8-bar sections. The three ``A'' sections are variations of each other. The last ``A'' section, shown in Figure~\ref{fig:strangers}, consists of a two-bar cell which is repeated three times. Each occurrence is a subtle variation of the preceding one. The second occurrence (bars 27-28) is a mere transposition of the original pattern by one descending tone. The third instance (bars 29-30) is also transposed, but with a slight modification in the melody, which creates a surprise and concludes the song. Bars 29-30 are both a variation of the original pattern in bars 25-26. Current models for music generation fail to reproduce such long-range similarities between musical patterns. In this example, it is statistically unlikely that bars 29-30 be almost identical to bars 25-26.

Our goal is to generate structured musical pieces from statistical models. Our approach is to impose a predefined musical structure that specifies explicitly repetitions and variations of patterns and sections, and use a statistical model to generate music that ``instantiates'' this structure. In this approach, musical structure is viewed as a procedural process, external to the statistical model.

An essential ingredient to implementing our approach is a mechanism to generate variation of a given musical pattern from a statistical model. Although it is impossible to characterise formally the notion of variation, it was shown that some measures of melodic similarity are efficient at detecting variations of a theme \cite{mongeau_sankoff_1990}. We propose to use such a similarity measure in a generative context to sample from a Markov model, patterns that are similar to a given pattern. This method is related to work on stochastic edit distances \cite{ristad1998learning,Cotterell2014}, but is integrated as a constraint in a more general model for the generation of musical sequences \cite{FlowComposerCP}. Moreover, our approach relies on an existing similarity measure rather than on labeled data (pairs of themes and related variations), which is not available.

This article is organised as follows. In Section~\ref{sec:melodic_similarity}, we present the Mongeau \& Sankoff similarity measure between melodies. In Section~\ref{sec:a_model_for_the_generation_of_variations}, we describe our model, based on melodic similarity, for the sampling of melodic variations. Section~\ref{sec:experimental_results} is an experimental evaluation of the variation sampling method. In Section~\ref{sec:examples_of_variations}, we show variations of a melody and a longer, structured musical piece.

\section{Melodic Similarity} 
\label{sec:melodic_similarity}

The traditional string edit distance considers three editing operations: \emph{substitution}, \emph{deletion}, and \emph{insertion} of a character. \cite{mongeau_sankoff_1990} add two operations motivated by the specificities of musical sequences, and inspired by the time compression and expansion operations considered in \emph{time warping}. The first operation, called \emph{fragmentation}, involves the replacement of one note by several, shorter notes. Similarly, the \emph{consolidation} operation, is the replacement of several notes by a single, longer note.

\newcommand{\w}[1]{\ensuremath{w_{\mathit{#1}}}}
\newcommand{\wdel}{\w{del}}
\newcommand{\wins}{\w{ins}}
\newcommand{\wfrag}{\w{frag}}
\newcommand{\wcons}{\w{cons}}
\newcommand{\wsub}{\w{subst}}
\newcommand{\wpitch}{\w{pitch}}
\newcommand{\wlen}{\w{len}}

Mongeau and Sankoff proposed an algorithm to compute the similarity between melodies in polynomial time.
Considering melodies as sequences of notes, let $A = a_1, \dots, a_m$ and $B=b_1, \dots, b_n$ be two melodies. The algorithm is based on dynamic programming, using the following recurrence equation for $i=1, \dots, m$ and $j=1, \dots, n$
\[
	\delta_{i,j} = \min{
		\begin{cases}
			\delta_{i-1,j} + \wdel(a_i) \\
			\delta_{i,j-1} + \wins(b_j) \\
			\delta_{i-1,j-1} + \wsub(a_i,b_j) \\
			\{\delta_{i-1,j-k} + \wfrag(a_i,b_{j-k+1},\dots,b_j), \forall k\} \\
			\{\delta_{i-k,j-1} + \wcons(a_{i-k+1},\dots,a_i,b_j), \forall k\}
		\end{cases}}
\]
where the various $w$ functions denotes predefined local weights associated to the basic editing operations.

The similarity between $A$ and $B$ is $\delta_{m,n}$, given the following initial conditions
\begin{itemize}
	\item $\delta_{i,0} = \delta_{i-1,0} + \wdel(a_i), i \geq 1$,
	\item $\delta_{0,j} = \delta_{0,j-1} + \wins(b_j), j \geq 1$,
	\item $\delta_{0,0} = 0$.
\end{itemize}

The weight of the substitution of a note $a_i$ with a note $b_j$ is the weighted sum of two weights:
\begin{equation}\label{eq:ms_subst}
	\wsub(a_i,b_j) = \wpitch(a_i,b_j) + k_1 \wlen(a_i,b_j),
\end{equation}
where $k_1$ is the predefined relative contribution of length difference versus that of pitch difference.

For the consolidation operation, the weight is also a weighted sum of \wpitch\ and \wlen, where \wlen\ is the difference between the length of the replaced note and the total length of the replacing notes, and where \wpitch\ is the sum of the \wpitch\ weights between each replacing note and the replaced note. The weight associated with the fragmentation operation is defined in the same manner.

The Mongeau \& Sankoff measure is well-adapted to the detection of variations, but has a minor weakness: there is no penalty associated with fragmenting a long note into several shorter notes of same pitch and same total duration. The same applies to consolidation. This is not adapted in a generative context, as fragmentation/consolidation change the resulting melody. We modify the original measure by adding a penalty $p$ to the weights of the consolidation and fragmentation operations:

The weight associated with a fragmentation of a note $a_i$ into a sequence of notes $b_{j-k+1},\dots,b_j$ is
\begin{align*}
 	\wfrag(a_i,b_{j-k+1},\dots,b_j)
 		&= \wpitch(a_i,b_{j-k+1},\dots,b_j) \\
 		&+ k_1 \len(a_i,b_{j-k+1},\dots,b_j) \\
		&+ p
\end{align*}
For consolidation, a similar extra-weight is added. The consolidation weight is defined by
\begin{align*}
 	\wcons(a_i,b_{j-k+1},\dots,b_j)
 	&= \wpitch(a_i,b_{j-k+1},\dots,b_j) \\
 	&+ k_1 \len(a_i,b_{j-k+1},\dots,b_j) \\
 	&+ p.
\end{align*}


\section{A Model for the Generation of Variations} 
\label{sec:a_model_for_the_generation_of_variations}

	Given an original theme, \ie a melody or a melody fragment, we generate variations of this theme by sampling a specific graphical model. This graphical model is a modified version of the general model of lead sheets introduced by~\cite{FlowComposerCP}. We first give a quick outline of this model, and then we show how we can adapt this model to restrict it to producing lead sheets at a controlled Mongeau \& Sankoff similarity measure from the theme.
	
	\subsection{The Model of Lead Sheets} 
	\label{sub:the_model_for_lead_sheet_generation}

\begin{figure}[htb!]
  \centering
    \scalebox{0.7}{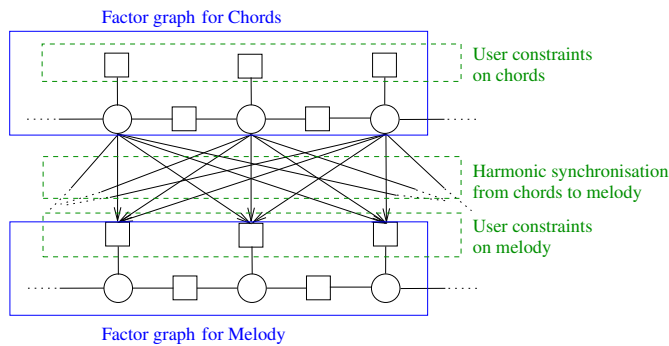}
  \caption{The two-voice model for lead sheet generation}
  \label{fig:full_model}
\end{figure}

	 \cite{FlowComposerCP} introduces a general, \emph{two-voice model} of lead sheets, which we briefly summarise. The overall model comprises \emph{two graphical models}, one for chord sequences, one for melodies. Both models are based on a factor graph that combines a Markov model with a finite state automaton. The Markov model, trained on a corpus of lead sheets, provides the stylistic model. The finite state automaton represents hard temporal constraints that the generated sequences should satisfy, such as metrical properties (\eg an imposed total duration) or user imposed temporal constraints. A \emph{harmonic model}, also trained on a corpus of lead sheets, is used to ``synchronise'' the chord sequences and the melody, ensuring that melodies contain notes that comply harmonically with the chord at the same temporal position. This temporal synchronisation is specified with the use of a temporal probability function $p_{\pi}$ attached to each model, \ie a conditional probability $p_{\pi}(e | t,e')$ of putting element $e$ after element $e'$, at temporal position $t$. Elements can be either chords or notes. For harmonic synchronisation, once a chord sequence is generated, one can define $p_{\pi}(e | t,e')$ to be the probability of placing note $e$ under the chord occurring at time $t$, and following a note $e'$. Conversely, an existing melody can be harmonised by generating a chord sequence in a similar fashion. This model is outlined on Figure~\ref{fig:full_model}.

		%

Each factor graph is made of a sequence of variables, represented with circles, encoding the sequence of elements. Each element is a chord or a note $e$, with a fixed duration $d(e)$. Unary factors, represented with a square linked to one variable, introduce a bias on its variable. This is used to encode unary constraints, \eg start the sequence with an imposed element. Binary factors, represented with a square linking two consecutive variables, combine the Markov transition probabilities with the finite-state automaton transitions, and are additionally used to incorporate to the model the temporal probabilities $p_{\pi}(e | t)$ used for harmonic synchronisation. The graphical model defines a distribution over the sequence of variables defined by the product of all unary and binary factors. A Belief Propagation algorithm samples the two models by taking into account partially filled fragments and propagating their effect to all empty sections. This is explained in detail in~\cite{FlowComposerCP}.


\subsection{Generating Variations on a Theme} 
\label{sub:generating_variations_on_a_theme}


	To generate variations of an imposed theme, we exploit the mechanism used for harmonic synchronisation. We modify the temporal probabilities $p_{\pi}$ by introducing a \emph{bias} on those temporal probabilities, \ie a bias $\beta(e|t,e')$ for putting element $e$ at time $t$ after $e'$. The new temporal probabilities become $p_{\pi}(e | t ,e').\beta(e|t,e')$, renormalised. As a result, the probability of a sequence in the new model is modified, compared with the original model, by a factor equal to the product of the biases $\beta(e|t,e')$, for all elements $e$ of the sequence.
	A bias of 1 does not modify the probability of putting element $e$ at time $t$ after $e'$, and a bias less than 1 decreases this probability. We set the value of $\beta(e|t,e')$ according to a ``localised'' similarity measure between the sequence $e',e$ and the fragment of the theme between $t-d(e')$ and $t+d(e)$.

		\begin{figure}[htb!]
			\begin{center}
				\includegraphics[width=\columnwidth]{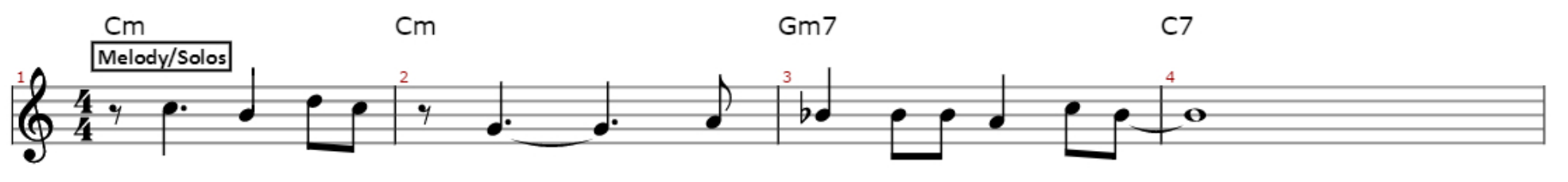}
			\end{center}
			\caption{The first four bars of \song{Solar}, by Miles Davis.}
			\label{fig:solar}
		\end{figure}

		The lead sheet in Figure~\ref{fig:solar} shows the first four bars of \song{Solar} by Miles Davis. Suppose we train a lead sheet model on a corpus of all songs by Miles Davis. Sampling this model would produce new lead sheets in the style of Miles Davis, but not necessarily similar to Solar specifically. However, it is possible to bias this general model to favour sequences with a \note{C}{5} dotted quarter note starting at beat 1.5 of the first bar, as in Solar, by setting $\beta(\note{C}{5} | t = 1.5,n') = 1$, for any preceding note $n'$, and by decreasing $\beta(n | t = 1.5,n')$ for all other note $n \neq \note{C}{5}$.
		
		We can also bias this model not to produce the same note at the same position, but to produce a similar musical result. For example, if we replace the \note{C}{5} dotted quarter note at beat 1.5 with a \note{C}{5} eighth note followed by a quarter note with the same pitch, as shown on Figure~\ref{subfig-2:mgd87}, we obtain a very similar musical output. On the contrary, replacing the \note{C}{5} dotted quarter note by a triplet of eighth notes \note{C}{5}-\note{B\flat}{4}-\note{C}{5} tied to a \note{C}{5} whole note, as shown on Figure~\ref{subfig-2:mgd906}, produces a totally different musical impression.

		Our approach to the generation of variations of a theme is to evaluate the similarity between each possible note $n$ at a given position $t$ and following note $n'$ in the generated sequence, and the notes in the theme around position $t$. We then set each bias $\beta(n | t,n')$ based on this similarity measure.

		Technically, for every candidate note $n$, we consider all potential temporal positions $t$ and all potential predecessors $n'$. We compute the Mongeau \& Sankoff similarity between the two-note melody $[n',n$] and the melodic fragment of the theme between time $t-d(n')$ and $t + d(n)$, where $d(n)$ is the duration of the note $n$, \ie the melodic fragment that would be replaced by placing the melody $[n',n]$, starting $n$ at time $t$. The notes of the theme that overlap the time interval $[t-d(n'),t + d(n)]$ are trimmed so that the extracted melody has the same duration as the candidate notes. We denote this distance $\text{MGD}([n',n],t)$. Similarly $\text{MGD}([n'],t)$ denotes the similarity of the one-note sequence $[n']$ starting at $t-d(n')$. Finally,  We refer to this similarity as the \emph{localised} Mongeau \& Sankoff similarity. The idea is that the similarity measure obtained by summing those localised measures over a complete sequence approximates the actual Mongeau \& Sankoff similarity. This will be confirmed experimentally in the next section.
		
		To convert the distance into a weight between 0 and 1, we rescale those values to the $[0,1]$ interval, and then invert their order, so that a value of 1 is the closest to the theme, and 0 the furthest away. Finally, we exponentiate the result, so that the logarithm of the product of the biases achieved by the model is proportional to the approximated Mongeau \& Sankoff similarity. Formally, we define $\beta(n | t,n')$ as follows, where $\text{MGD}_{\mathit{max}}$ is the maximal value of localised Mongeau \& Sankoff similarities:
		\[
		\beta(n | t,n') = \exp\left( 1 - \frac{\text{MGD}([n',n],t) - \text{MGD}([n'],t)}{ \text{MGD}_{\mathit{max}} } \right)
		\]
		An additional parameter $\alpha$ is used to control the strength of the variation mechanism. This parameters affects the slope of the set of values taken by the bias, by redefining $\beta'(n | t,n') = (1-\alpha).\beta(n | t,n') + \alpha$. When $\alpha$ is 0, this does not modify the bias, and as $\alpha$ increasing to 1, the bias decrease in strength, and with a value of 1 there is not bias in favour of the theme at all, and the modified model is equivalent to the original one. In between values have an intermediate effect, and decrease the bias less quickly for notes that are further to the theme.
		
		Consequently, small values of $\alpha$ should lead to melodies that are highly similar to the theme. As $\alpha$ increases, the generated melodies should be less imitative of the original theme. We will evaluate this in practice in the next section.



\section{Experimental Results} 
\label{sec:experimental_results}

	The approach we proposed is based on the intuition that \emph{local} similarities, favoured by the biased model, will result in a \emph{global} similarity between the generated melodies and the theme. In this section, we evaluate our approach according to two aspects. First, we evaluate how the choice of the value for the parameter $\alpha$ influences the Mongeau \& Sankoff similarity between the generated melodies and the original theme. In particular, we show that the biased model favours sequences closer to the theme and penalises sequences less similar to the theme. Second, we explain the result more analytically, for $\alpha = 0$. We first show that applying the bias to the model approximates the localised Mongeau \& Sankoff similarity, and then we show that this localised Mongeau \& Sankoff similarity is a good approximation of the actual Mongeau \& Sankoff similarity.
	
	For the experiments reported in this section, the original theme consists in the melody in the first four bars of \song{Solar} (Miles Davis, see Figure~\ref{fig:solar}). The training corpus contains 29 lead sheets by Miles Davis, such as \song{All Blues}, \song{Nardis}, \song{So What} or \song{Solar} itself. In each experimental setup, we build a general model of 4-bar lead sheets in the style of Miles Davis, called the \emph{unbiased model}, and then, the model is biased to favour the theme with some value for $\alpha$, resulting in a \emph{biased model}.

	\subsection{Correlation between the Biases and the Mongeau \& Sankoff Distance} 
	\label{sub:correlation}
	
\begin{figure}[htb!]
			\begin{center}
				\includegraphics[width=0.8\columnwidth]{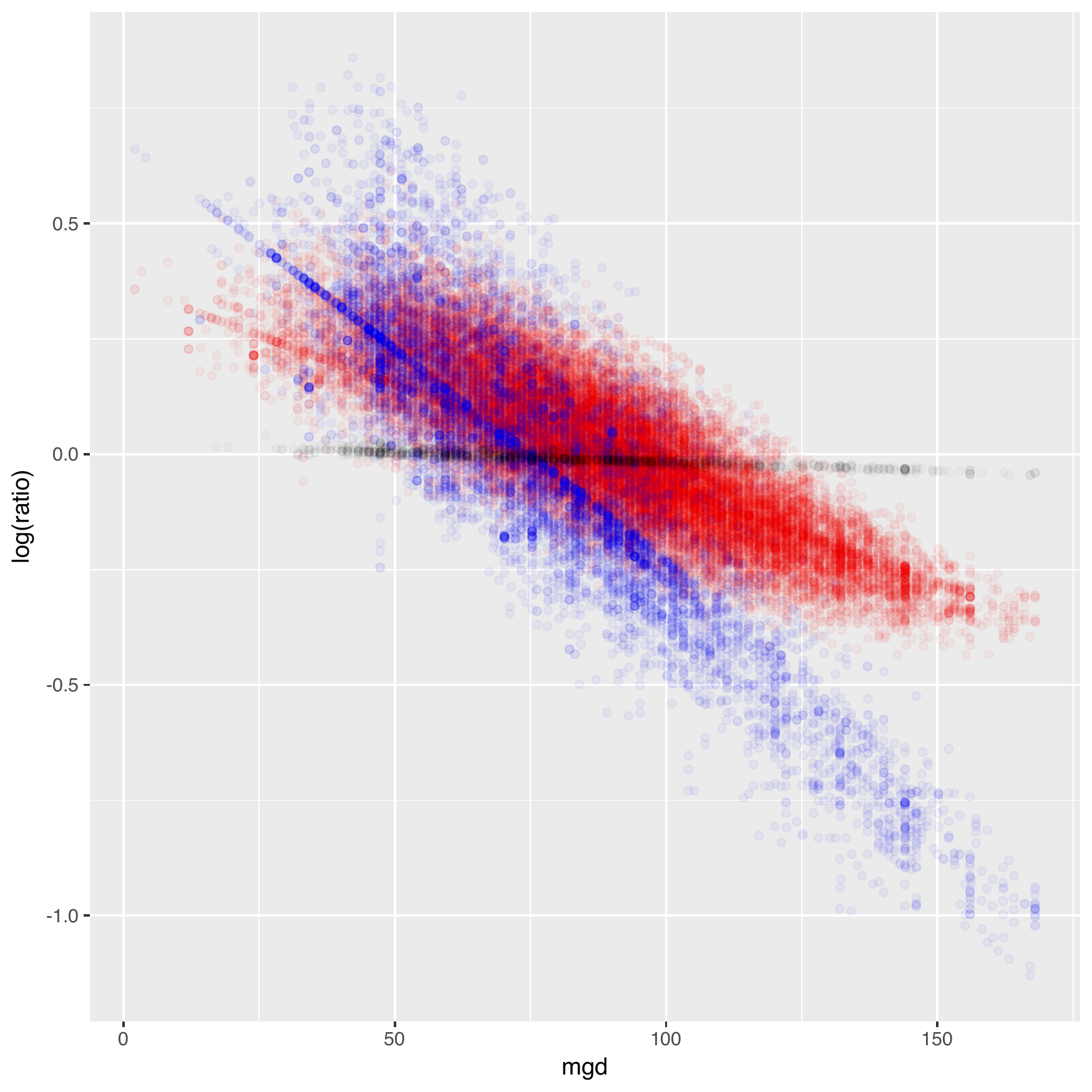}
			\end{center}
			\caption{Sequence probability ratio (log) against Mongeau \& Sankoff similarity to theme. Sequences in blue, red and black have been generated with $\alpha = 0, \alpha = 0.5, \alpha = 0.95$, respectively.}
			\label{fig:general_results}
		\end{figure}

	For one value of $\alpha$, we generate 10 000 variations of the original theme (first four bars of \song{Solar}). For each sequence, we compute its probability $p_o$ in the unbiased model and its probability $p_b$ in the biased model, and then consider the ratio $p_b/p_o$. This probability ratio shows by how much the sequence has been favoured, for values greater than 1, or conversely penalised, for values less than 1, in the biased model. On Figure~\ref{fig:general_results}, points in blue are sequences generated with the biased model with an $\alpha=0$, \ie the strictest possible. For each sequence, we plot its probability ratio, on a log scale, against its Mongeau \& Sankoff similarity with the theme. We observe that the logarithm of the probability ratio tends to decrease linearly as the Mongeau \& Sankoff distance with the theme increases. Sequences at a distance less than 75 from the theme are boosted while sequences at a distance more than 75 from the theme are hindered. Points in black are sequences generated with $\alpha=0.95$, \ie almost no bias at all. We observe that most sequences have a probability ratio of 1, \ie that the biased model hardly affects the probability of sequences. Only sequences very far from the theme have their probability slightly decreased. Points in the red are generated with $\alpha=0.5$. They display an intermediate behaviour as expected.
	
	Examples of variations of the first four bars of \song{Solar}, at various Mongeau \& Sankoff distances, are shown in Section~\ref{sub:first_four_bars_of_solar}.


	\subsection{Explaining the Correlation} 
	\label{sub:explaining_the_correlation}
	
	We explain the correlation observed by the application of two successive approximations. We concentrate on the case where $\alpha=0$, but similar results are obtained with other values. We can break our analysis in three steps.
	
	First, we note that for a given sequence, its probability ratio is equal, by definition of the biased model, to the product of all the local biases applied to each element of the sequence, up to a normalisation factor. This has been verified experimentally too, although we do not show the plot for reasons of space.
	
\begin{figure}[htb!]
			\begin{center}
				\includegraphics[width=0.8\columnwidth]{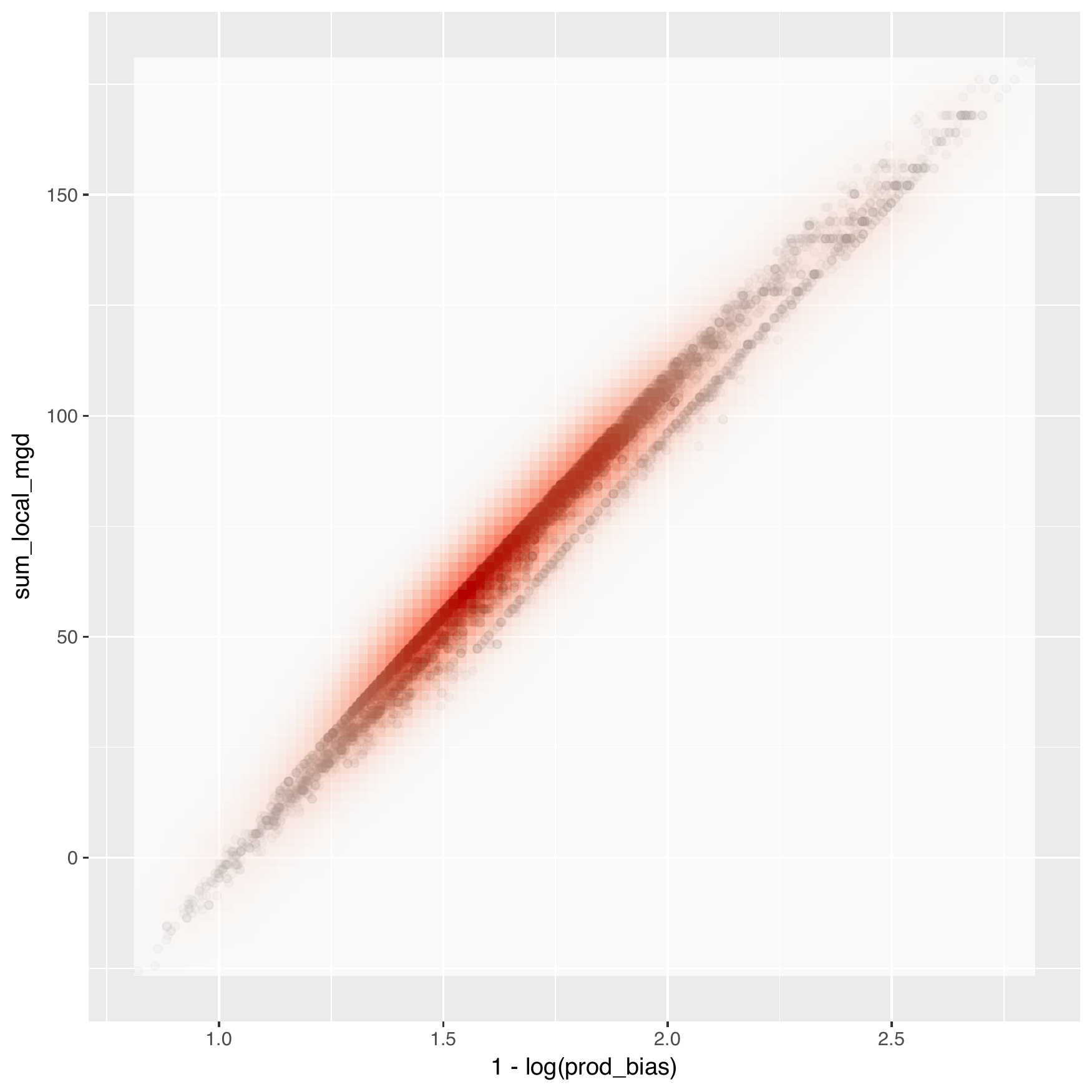}
			\end{center}
			\caption{The sum of localised Mongeau \& Sankoff similarity measures against the product of local biases (log), for $\alpha=0$}
			\label{fig:localised_prod_bias}
		\end{figure}

	Second, we show how the probability ratio compares with the approximated Mongeau \& Sankoff similarity measure obtained by summing the localised Mongeau \& Sankoff similarity measures. For each sequence, we sum, over all its elements, the localised Mongeau \& Sankoff that was used when computing the biases, as explained in section Section~\ref{sub:generating_variations_on_a_theme}. Then, we compare this sum to the product of the local biases, equal to the probability ratio. We plot the result on Figure~\ref{fig:localised_prod_bias}. We observe that the approximated Mongeau \& Sankoff similarity measure is tightly correlated with the logarithm of the product of the local biases. In other words, the logarithm of the product of the local biases approximates closely enough the sum of the localised Mongeau \& Sankoff distances.
	
\begin{figure}[htb!]
			\begin{center}
				\includegraphics[width=0.8\columnwidth]{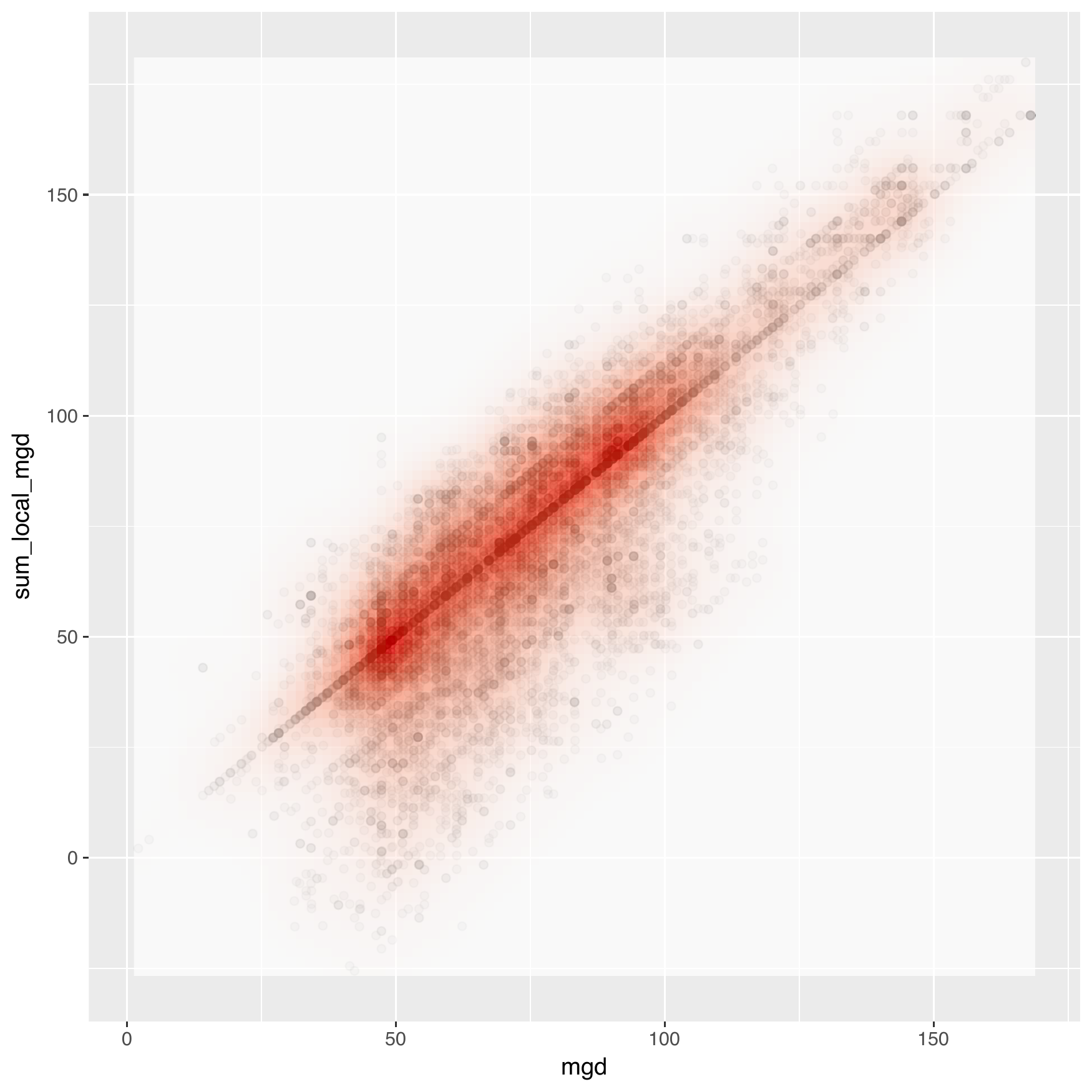}
			\end{center}
			\caption{The sum of localised Mongeau \& Sankoff similarity measures against the actual Mongeau \& Sankoff measure, for $\alpha=0$}
			\label{fig:localised_mgd}
		\end{figure}

	Finally, we show that this approximated Mongeau \& Sankoff similarity measure approximates the actual Mongeau \& Sankoff similarity measure. On Figure~\ref{fig:localised_mgd}, we plot for each sequence, the approximate versus the actual similarity measure. We observe that, although the actual measure is a global, dynamic programming-based measure, it is adequately approximated by summing the localised versions. This is probably because the localised measure captures sufficiently the effect of a note on the global similarity measure.

\section{Examples of Variations} 
\label{sec:examples_of_variations}

In this section, we show music created using the variation sampling mechanism. We show melodic variations of a short melodic fragment from \song{Solar} (Section~\ref{sub:first_four_bars_of_solar}) and of the whole melody (Section~\ref{sub:variations_of_entire_lead_sheets}). Section~\ref{sub:structured_lead_sheets} shows a 32-bar lead sheet that has the same structure as \song{In a Sentimental Mood} by Duke Ellington, but in a very different style (the Beatles).

	\subsection{First Four Bars of \song{Solar}} 
	\label{sub:first_four_bars_of_solar}

\begin{figure}[htb!]
		\subfloat[Mongeau \& Sankoff distance 12: highly similar to the theme\label{subfig-1:mgd12}]{%
			\includegraphics[width=\columnwidth]{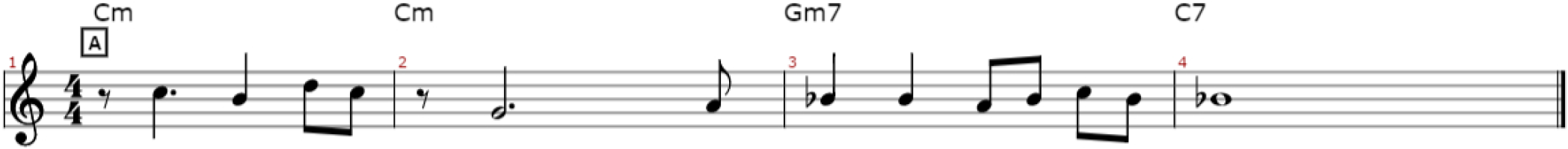}
		}

		\subfloat[Distance 86, minor enrichments in bars 1 and 3\label{subfig-2:mgd86}]{%
			\includegraphics[width=\columnwidth]{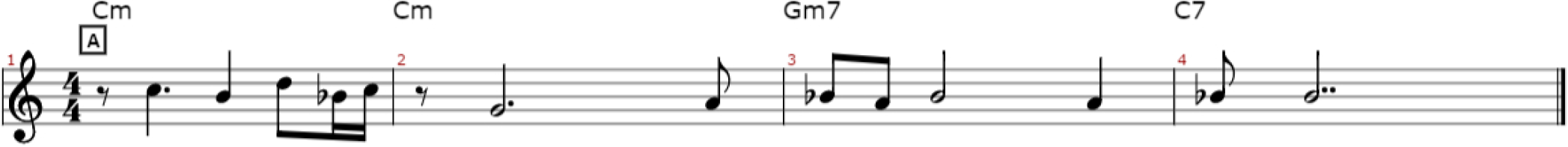}
		}

		\subfloat[Distance 87, minor enrichments in bars 1 and 3\label{subfig-2:mgd87}]{%
			\includegraphics[width=\columnwidth]{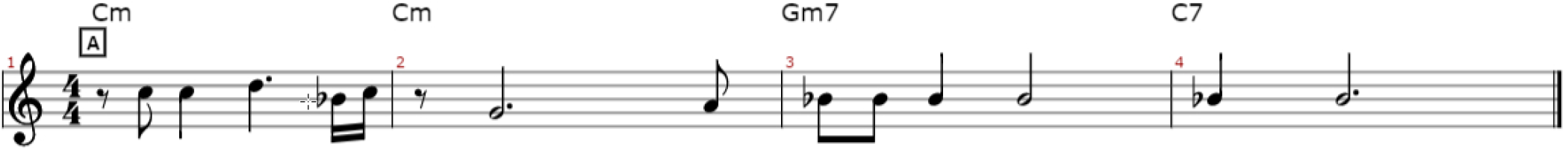}
		}

		\subfloat[Distance 224, with major differences in bars 2 and 3\label{subfig-2:mgd224}]{%
			\includegraphics[width=\columnwidth]{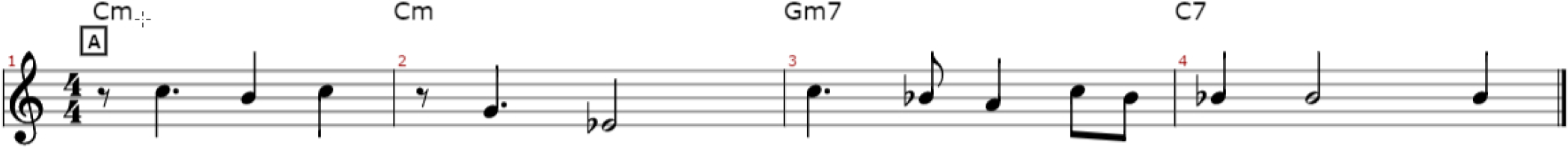}
		}

		\subfloat[Distance 285, interesting triplet rhythm in bar 1\label{subfig-2:mgd285}]{%
			\includegraphics[width=\columnwidth]{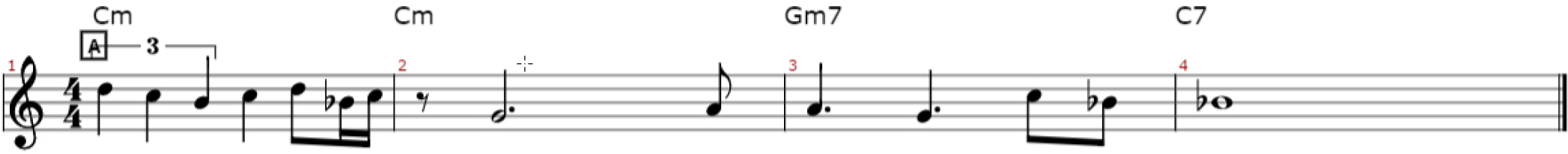}
		}

		\subfloat[Dist. 295, large initial interval (octave) and end of bar 3 differs from other variations\label{subfig-2:mgd295}]{%
			\includegraphics[width=\columnwidth]{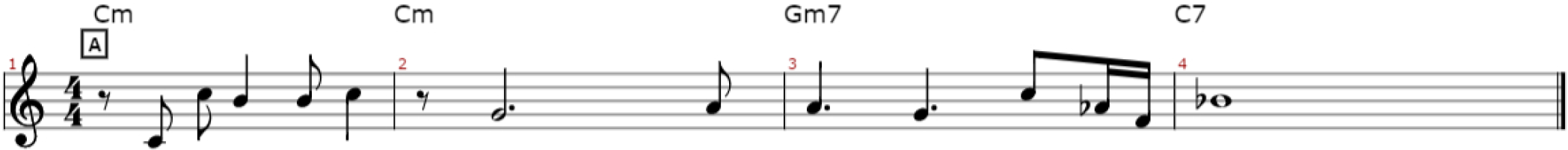}
		}

		\subfloat[Dist. 906, first bar uses a rhythm similar that of \song{Miles Ahead} (Miles Davis), and bar 3 is introduces a new rhythm, similar to that of the original theme, except with dotted quarter notes\label{subfig-2:mgd906}]{%
			\includegraphics[width=\columnwidth]{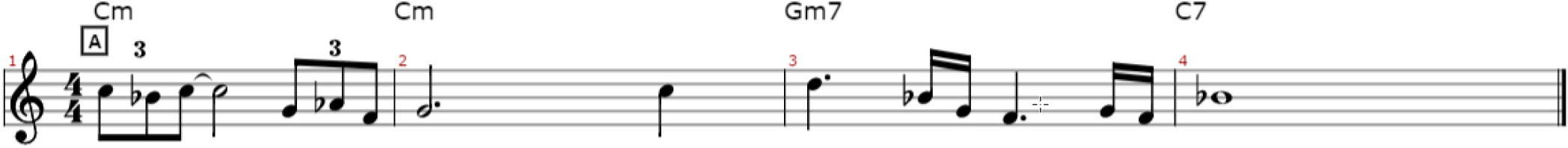}
		}
		\caption{Several variations of the first four bars of \song{Solar}, by increasing Mongeau \& Sankoff distance.}
		\label{fig:solar_variations}
	\end{figure}

	Figure~\ref{fig:solar_variations} shows six melodic variations of the first four bars of \song{Solar}, by Miles Davis. These variations were created using a model trained on 29 songs by Miles Davis (see Section~\ref{sec:experimental_results}). The variations are presented in increasing order of Mongeau \& Sankoff distance with the original theme (see Figure~\ref{fig:solar}). Note that the variations are increasingly different from the theme, both rhythmically and melodically.

	\subsection{Variations of Entire Lead Sheets} 
	\label{sub:variations_of_entire_lead_sheets}

\begin{figure}[htp!]
		\begin{center}
			\includegraphics[width=\columnwidth]{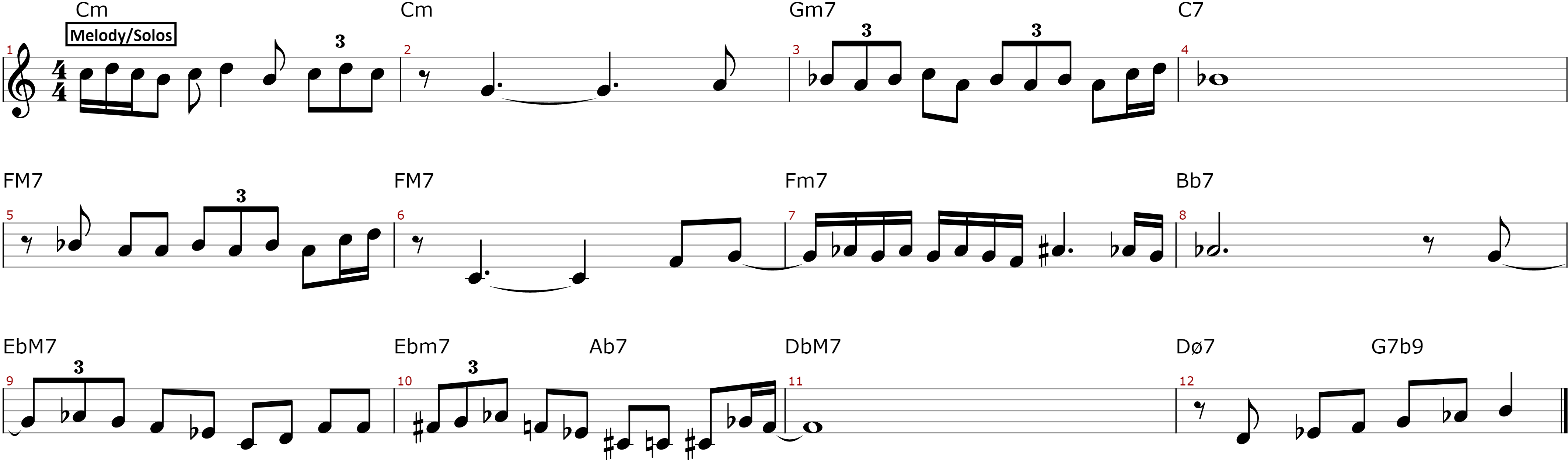}
		\end{center}
		\caption{This is a variation of Solar with the original chord labels and with the constraint that the variations should contain more notes (here, 77 notes including rests) than the original theme (48 notes including rests). Note that bars 9-10 are melodically interesting and are similar to each other, very much like in the original theme.}
		\label{fig:solar_var_more_notes}
	\end{figure}

	Figure~\ref{fig:solar_var_more_notes} shows a variation of \song{Solar}, in the style of Miles Davis (the system was trained using the same corpus as in Section~\ref{sec:experimental_results}). This variation contains a lot more notes that the original melody, \ie 77 notes instead of 48 notes, including rests and many triplets. Bars 2, 4, 6, 8, and 11-12 are unchanged from the original melody. Bars 9-10 are similar to one another, very much like in the original theme, but are more complex than the original bars. This variation feels like an ornamented version of \song{Solar}, that is a \song{Solar} ``with more notes''.

\begin{figure}[htp!]
		\begin{center}
			\includegraphics[width=\columnwidth]{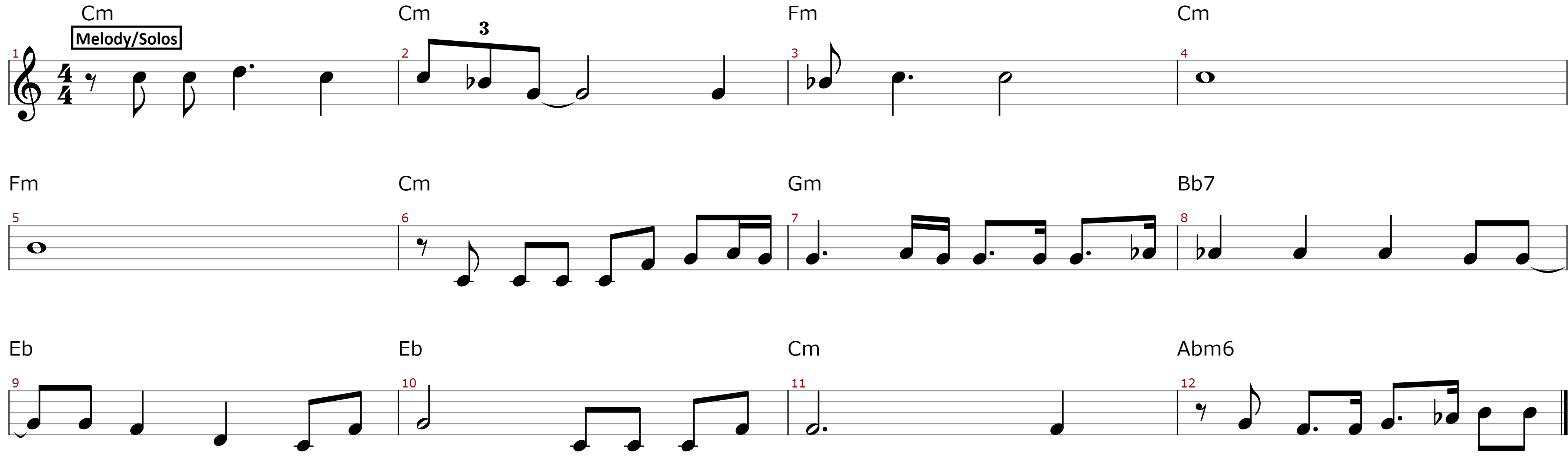}
		\end{center}
		\caption{This is a harmonic and melodic variation of \song{Solar} in the style of the Beatles. The chord sequence is a variation of the original chord sequence of Solar, using a distance between chord labels.}
		\label{fig:solar_var_harmo_melody_beatles}
	\end{figure}

	The mechanism described in this chapter creates variations based on the Mongeau-Sankoff distance between melodies. This mechanism may be applied to other types of sequences and other distances, for instance to chord sequences, using any harmonic distance between chords. The lead sheet on Figure~\ref{fig:solar_var_harmo_melody_beatles} was generated in two steps: first, a 12-bar chord sequence was generated as a variation of the chord sequence of \song{Solar}, then a melodic variation was generated from the original melody. The similarity between chords is computed as the scalar product of the pitch hitograms of the chords. Figure~\ref{fig:solar_var_harmo_melody_beatles} has been generated in the style of the Beatles, using a corpus with 45 lead sheets by the Beatles.

	The lead sheets in Figures~\ref{fig:solar_var_more_notes}~and~\ref{fig:solar_var_harmo_melody_beatles} were played to several musical experts, who rated them as pleasing and not lacking a sense of direction. This supports the intuition that variations retain some of the structural features of the original pieces. These two lead sheets are not original compositions as they are created by transforming an existing piece. In the next section, we will show how the variation mechanism may be used to generate new, structured compositions.


	\subsection{Composition of Structured Lead Sheets} 
	\label{sub:structured_lead_sheets}

	\begin{figure}[htb!]
		\begin{center}
			\includegraphics[width=\columnwidth]{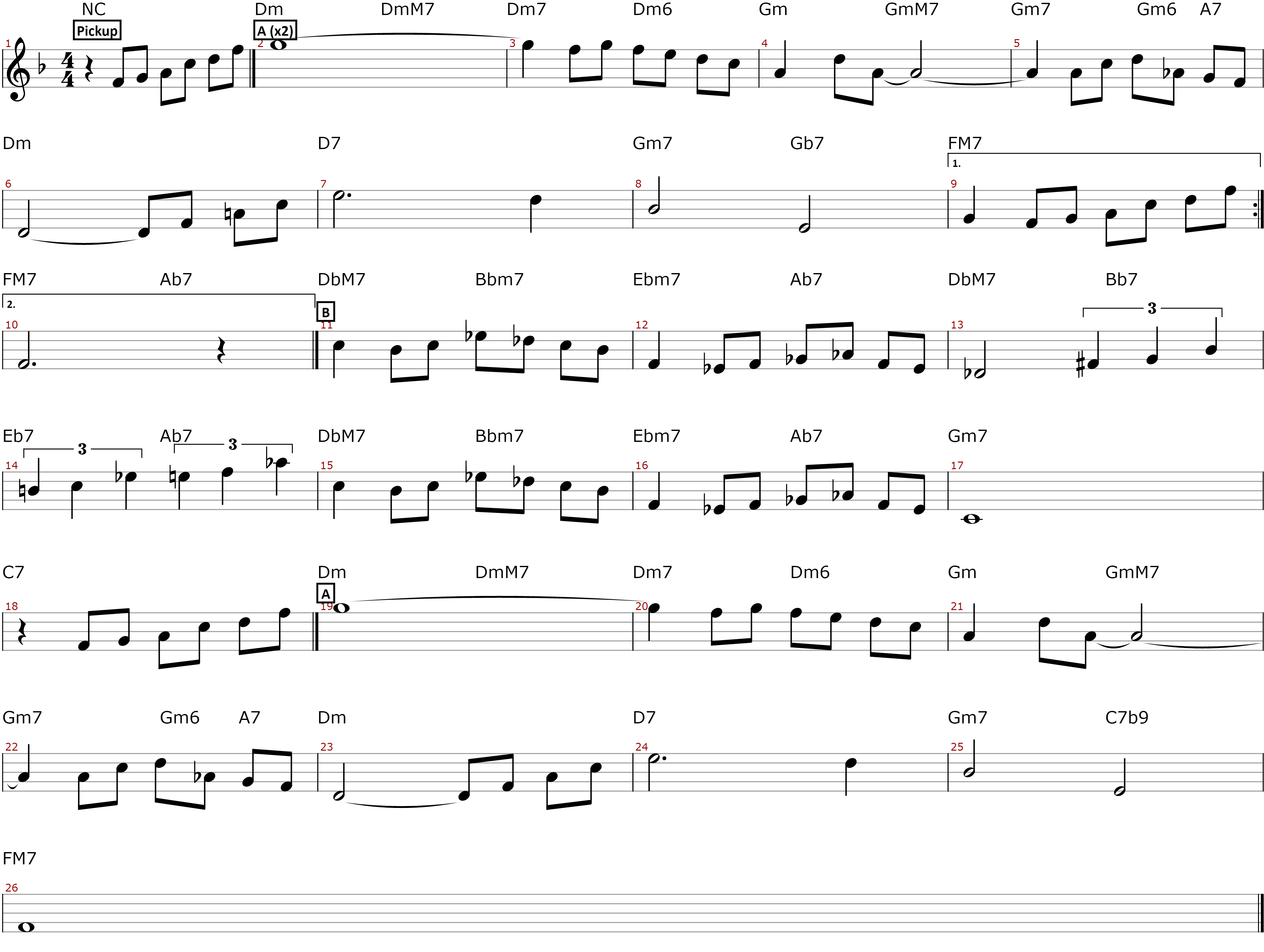}
		\end{center}
		\caption{\iasm\ by Duke Ellington is a typical 32-bar AABA song, with a pickup bar. Note that bars 11-12 are transposed variations of each other, that bars 15-16 are repetitions of bars 11-12, and the ending is a variation of the ending of the first ``A'' section.}
		\label{fig:iasm_lsdb}
	\end{figure}

	As said in introduction, the model of melodic variations is an essential tool for the generation of structured lead sheets. We illustrate this on a concrete example: generating a lead sheet with the structure of \iasm\ (Duke Ellington, see Figure~\ref{fig:iasm_lsdb}). This song is a classical AABA 32-bar form preceded by a pickup bar. More precisely, it consists of the following elements:
	\begin{itemize}
		\item Sections: ``pickup'': bar 1; ``A1'': bars 2-9, ``A2'': bars 2-8 \& 10 (bars 1-8 are played twice), ``B'': bars 11-18; ``A3''
		\item Bars 2-3 use similar chords (all based on \ch{Dm}) and the harmony is transposed down by a perfect fifth at bars 4;
		\item Bar 12 is a transposed variation of bar 11;
		\item Bars 15-16 are exact copies of bars 11-12;
		\item Last two bars (25-26) are a variation of the ending of Section ``A2'' (\ie bars 8 and 10)
	\end{itemize}

	The generation is based on a procedure that creates patterns, and variations, and organizes them to comply with an imposed structure. This procedure uses the general two-voice model of lead sheets \cite{FlowComposerCP} to generate original patterns and the model of variations of Section~\ref{sec:a_model_for_the_generation_of_variations} to generate variations of these patterns.

	In practice, the generation follows a left-to-right order, as much as possible. However, repetitions break the left-to-right scheme. For instance, bar 2 is played three times, once at the beginning of each ``A'' section. Bars that are copied from the past are later integrated as \emph{constraints} (as explained Section~\ref{sec:a_model_for_the_generation_of_variations}) in the model when generating the surrounding elements. For instance, bar 19 at the beginning of Section ``A3'', which is a copy of bar 2, will be treated as a constraint when generating the music for the preceding bars (end of Section ``B'').

	\begin{figure}[htb!]
		\begin{center}
			\includegraphics[width=\columnwidth]{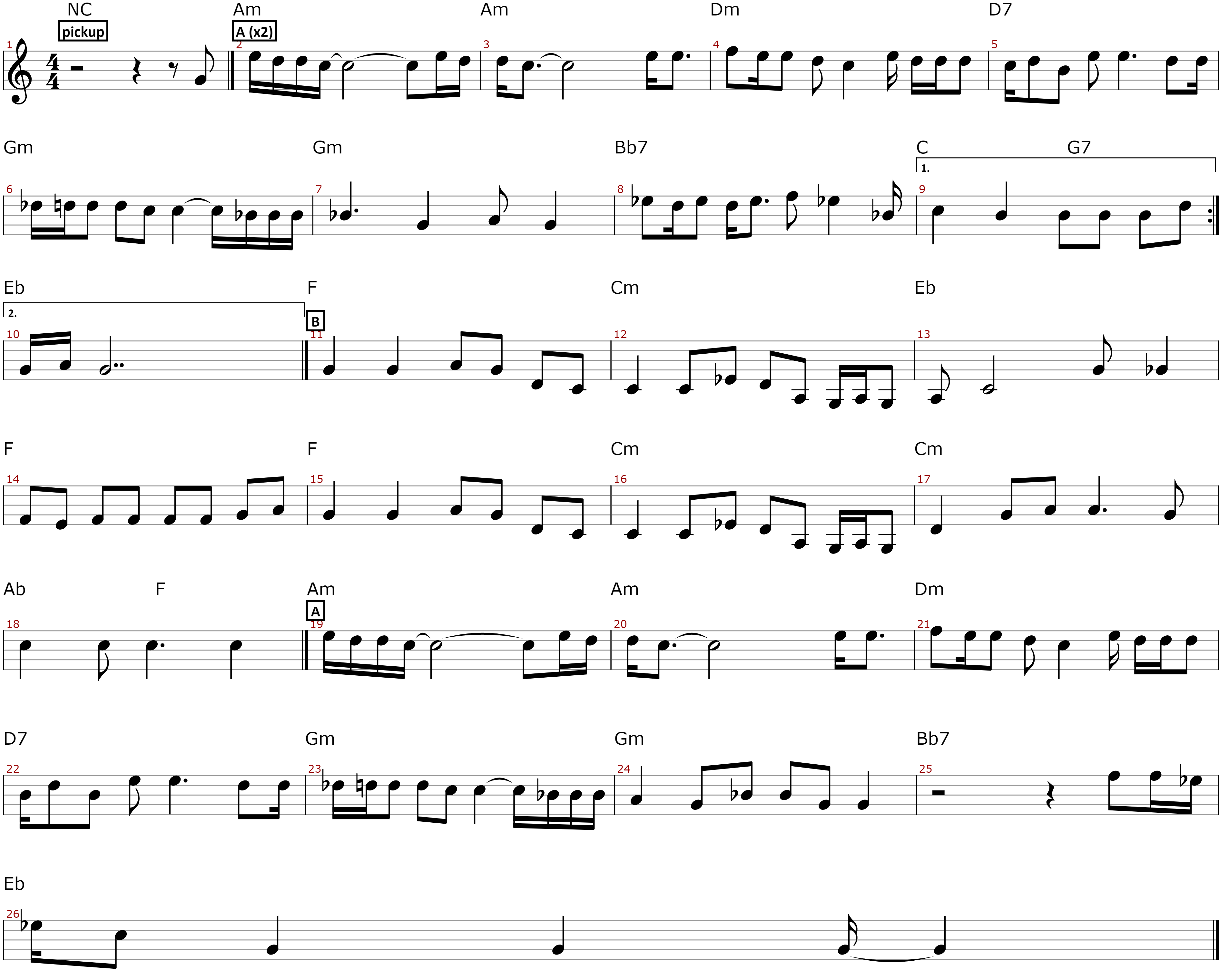}
		\end{center}
		\caption{A lead sheet with the structure of \song{In a Sentimental Mood} but in the style of the Beatles. Note that bar 12 is a transposed variation of bar 11, as in the original song. The ending is also a variation of the ending of Section ``A1''.}
		\label{fig:iasm_beatles}
	\end{figure}
	
	Figure~\ref{fig:iasm_beatles} shows a lead sheet with this structure and generated from a stylistic model of the Beatles (trained from a corpus with 201 lead sheets by the Beatles). The music does not sound similar to \iasm\ at all, but its structure, with multiple occurrences of similar patterns, make it feel like it was composed with some intentions. This is never the case of structureless 32-bar songs composed from the general model.

	Each part of the lead sheet has a strong internal coherence. The melody in the ``A'' parts use mostly small steps and fast sixteenth notes, many occurrence of a rhythmic pattern combining a sixteenth note with a dotted eighth note. The ``B'' part uses many leaps (thirds, fourth and fifth) and a regular eighth note rhythm. This internal coherence is a product of the imposed structure. For instance, in the ``B'' part, four out of eight bars come from a single original cell, consisting of bar 11.

	The fact that the ``A'' and ``B'' parts contrast with one another is also a nice feature of this lead sheet. This contrast simply results from the default behavior of the general model of lead sheets.



\section{Conclusion} 
\label{sec:Conclusion}

We have presented a model for sampling variations of melodies from a graphical model. This model is based on the melodic similarity measure proposed by \cite{mongeau_sankoff_1990}. Technically, we use an approximated version of the Mongeau \& Sankoff similarity measure to bias a more general model for the generation of music. Experimental evaluation shows that this approximation allows us to bias the model towards the generation of melodies that are similar to the imposed theme. Moreover, the intensity of the bias may be adjusted to control the similarity between the theme and the variations. This makes this approach a powerful tool for the creation of pieces complying with an imposed musical structure. We have illustrated our method on melodic variations of a well-known theme and we have shown that this is a building block for a procedure that generates \emph{structured} musical pieces.


\bibliographystyle{plain}
\bibliography{variations}

\end{document}